\documentclass{article} 
\usepackage{aaai2026}  
\usepackage{times}  
\usepackage{helvet}  
\usepackage{courier}  
\usepackage[hyphens]{url}  
\usepackage{graphicx} 
\usepackage[numbers,sort&compress]{natbib}  
\usepackage{caption} 
\usepackage{algorithm}
\usepackage{algorithmic}
\usepackage{pdfpages}
\usepackage{float}
\usepackage{multirow}
\usepackage{pifont}
\usepackage{algorithmic}
\usepackage{algorithm}   
\usepackage{subcaption}
\usepackage{multirow}
\usepackage{makecell}
\usepackage{booktabs}
\usepackage{amsmath}  
\usepackage{amssymb}
\usepackage{caption}
\usepackage[capitalize]{cleveref}
\usepackage{bibentry}

\usepackage{newfloat}
\usepackage{listings}
\DeclareCaptionStyle{ruled}{labelfont=normalfont,labelsep=colon,strut=off} 
\floatstyle{ruled}
\newfloat{listing}{tb}{lst}{}
\floatname{listing}{Listing}
\title{Consistent and Editable: A Balanced Framework for\\
Text-Guided Video Editing}
\author{
Tao Jin\textsuperscript{\rm 1}\thanks{First author. E-mail: jt0618@mail.ustc.edu.cn},
Li Xiao\textsuperscript{\rm 1}\thanks{Corresponding author. E-mail: xiaoli11@ustc.edu.cn}
}
\affiliations{
\textsuperscript{\rm 1}
 University of Science and Technology of China
}
\date{}

\begin{document}
\makeatletter
\let\copyright@text\relax
\let\@copyrightspace\relax
\makeatother

\maketitle

\begin{abstract}
Recently, diffusion models have achieved considerable success in the text-guided video editing domain. However, existing works often struggle to balance the trade-off between temporal consistency and editability in video editing, with consistency and editability typically being inversely related. To address this, we propose a high-quality video editing framework enhanced for consistency and editability, named EquiEdit, which improves coordinatively the temporal consistency and editability of the edited videos while achieving a balance between the two.
In terms of temporal consistency, the proposed temporal Mamba module with a tailored temporal-aware scanning scans fused video sequences following four designed directions, effectively enhancing the inter-frame consistency of edited videos. For editability, we design a noise injection strategy based on the spectral transformation to increase editing flexibility, where the Fourier transform is used to preserve the hidden structure in the initial latent noise used for editing, ensuring inter-frame consistency of the edited video and fidelity to the input video. Extensive qualitative and quantitative experiments demonstrate the effectiveness of our method in terms of temporal consistency and editability, as well as its great fidelity to the input video itself.
\end{abstract}


\section{Introduction}
\label{sec:intro}

As a cutting-edge generative model in the AIGC wave, diffusion models have sparked rapid development in multiple tasks related to images, including image generation~\cite{kumagai2023story,kingma2021variational,liu2023more}, image translation~\cite{parmar2023zero,li2022vqbb}, super-resolution~\cite{wu2024diffusion,saharia2022image}, and image editing~\cite{mao2024mag,avrahami2023blended,meng2021sdedit}. In image editing tasks guided by text prompts, diffusion-based methods~\cite{saharia2022photorealistic, gal2022image} achieve diverse and high-quality results due to their powerful controllability, stability, and amazing realism. This natural language-guided approach not only opens up a new paradigm for image editing, but also provides the possibility for ordinary users, even those without computer expertise, to freely engage in image editing. 
Success in text-to-image (T2I) editing paves the way for text-to-video (T2V) development.

\begin{figure}[t]
    \centering
    \captionsetup{type=figure}
    \includegraphics[width=0.48\textwidth]{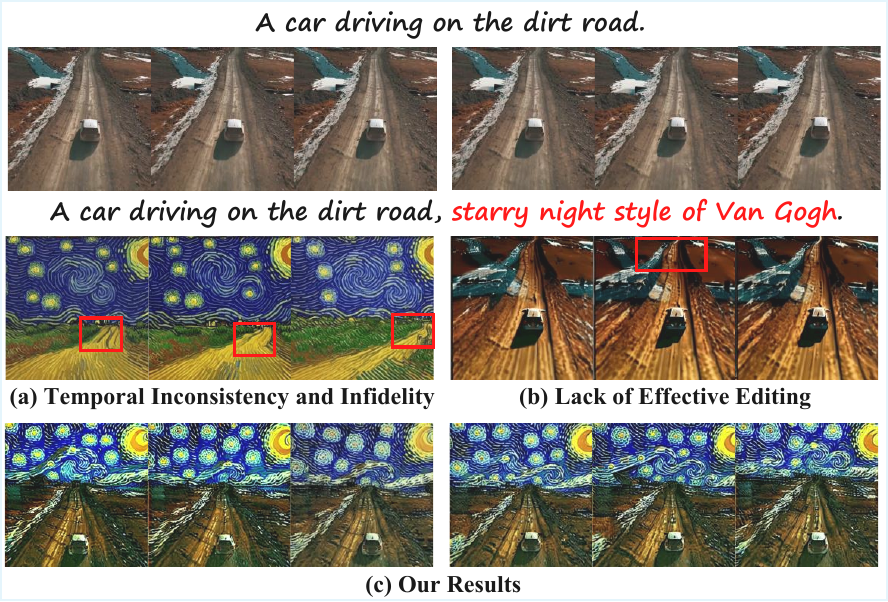}
    \caption{Illustration of temporal inconsistency and infidelity to the input video (a) and lack of effective editing (b). For comparison, we present our results in (c).}
    \label{fig:pdf_obseration} 
\end{figure}

In the era of short video, compared with static text and images, videos can present dynamic information and are the dominant force on the Internet~\cite{xing2023survey}. Therefore, video editing is of great significance to the media and entertainment industry.
However, unlike static image editing, maintaining coherence and temporal consistency between output video frames is one of the main challenges faced by video editing. To solve it, some methods~\cite{feng2024ccedit,wang2023edit,chen2023control} of training a T2V model on large-scale text-video pairs datasets can maintain consistency between output video frames, but they are time-consuming and computationally expensive, and obtaining large-scale text-video datasets such as WebVid-10M~\cite{bain2021frozen} is difficult. On the other hand, videos edited using fine-tuning approaches~\cite{zhang2024camel} that introduce temporal modules into pre-trained T2I models often exhibit temporal inconsistencies, such as flickering, lack of fidelity to the input video. In Fig.~\ref{fig:pdf_obseration} (a), although the edited result from SimDA~\cite{xing2024simda} closely matches the ``starry night style of Van Gogh", it demonstrates inter-frame inconsistencies and infidelity to the input video. In Fig.~\ref{fig:pdf_obseration} (b), while being faithful to the input video, the edited result from CAMEL~\cite{zhang2024camel} lacks effective editing. In summary, these methods usually lead to a trade-off between temporal consistency and editability.

\begin{figure*}[t]
    \centering
    \includegraphics[width=\textwidth]{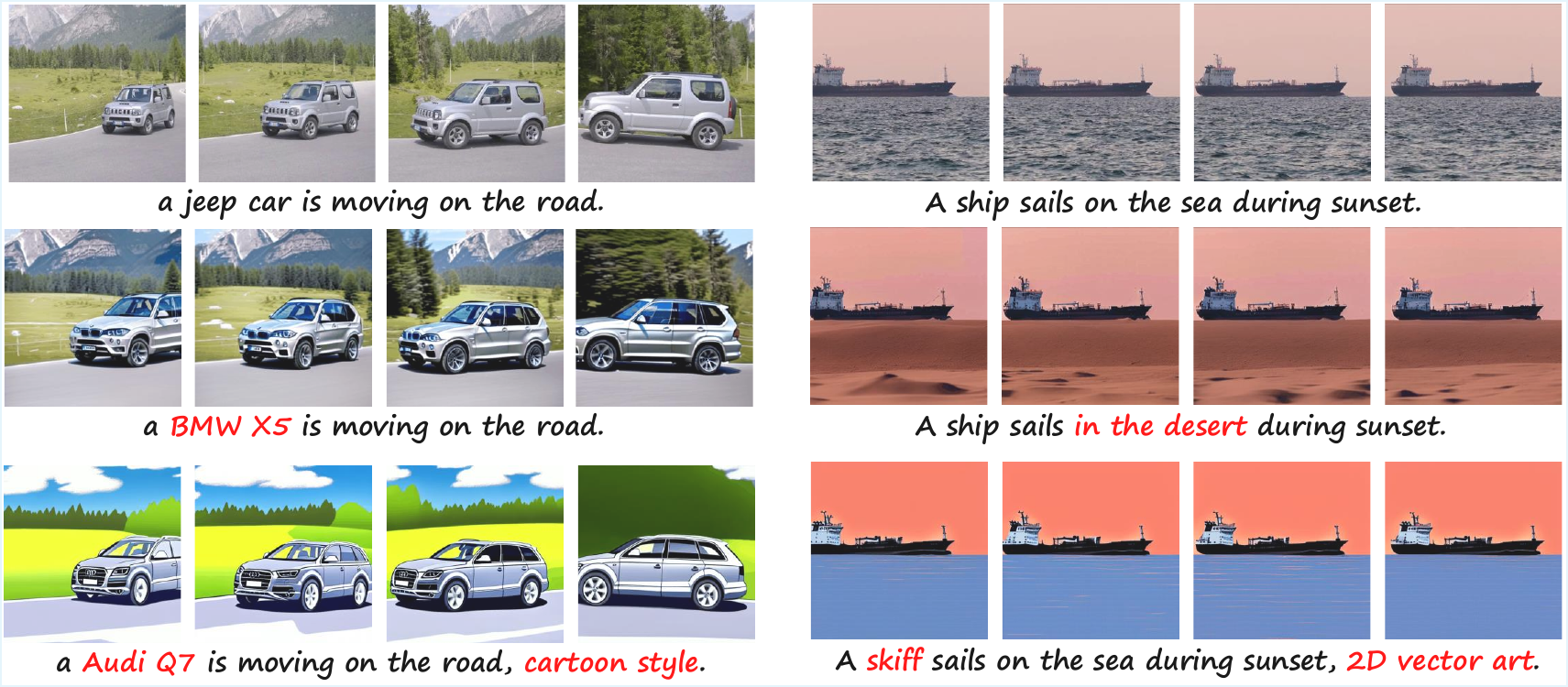}
    \caption{EquiEdit can improve and balance temporal consistency and editability, ensuring seamless and natural edits.}
    \label{fig:pdf_teaser} 
\end{figure*}

Although the aforementioned methods have achieved good editing results, balancing temporal consistency and editability remains an open problem. To address it, we attempt to promote video editing performance from the following two aspects: (1) \textbf{How to effectively enhance temporal consistency of video editing.} We designed a Mamba-based~\cite{gu2023mamba} module with temporal-aware scanning to effectively enhance the temporal consistency in the edited video. (2) \textbf{How to enhance video editability while maintaining temporal consistency.} A noise injection strategy is proposed to improve the flexibility of the diffusion model during the sampling, while utilizing spectral modulation to stabilize the coherence of the video. 

Recently, state space models (SSMs)~\cite{gu2021efficiently}, as the cornerstone of Mamba, have gained widespread attention due to their computational efficiency in modeling long sequences. 
However, the potential of Mamba in temporal modeling for video editing has yet to be fully explored; meanwhile, videos are often long sequences and Mamba tends to forget in long-context modeling~\cite{ye2025longmamba}. Inspired by the application of Mamba in the visual field, we propose a temporal Mamba high-quality video editing framework enhanced for consistency and editability, namely EquiEdit, which is fine-tuned in a one-shot manner~\cite{wu2023tune} based on a pre-trained T2I model. Unlike most methods that directly scan visual token sequences~\cite{zhu2024vision,li2024videomamba}, we propose the temporal Mamba module with designed temporal-aware scanning, which integrates visual and temporal information to construct a fused sequence. By scanning this fused sequence in the designed four spatial-first directions, we ultimately enhance the temporal consistency of edited videos.
For enhancing editability, we propose a noise injection strategy based on the Fourier transform to refine the initial latent noise for the sampling process.
At the same time, the flickering and blurring of the current video editing systems are caused by a high-frequency leak during the denoising process~\cite{yoon2024frag}. Thus, utilizing the Fourier transform, we preserve and enhance the structure-corresponding parts in the latent noise used for editing, which avoids a decrease in consistency and fidelity to the input. 
Despite existing studies on noise introduction~\cite{yoon2024dni},  our strategy differs in that it more cautiously controls the noise injection to ensure semantic consistency and fidelity.
In summary, our main contributions are as follows: 
\begin{itemize}
\item[$\bullet$] We propose EquiEdit, a novel, one-shot video editing framework. EquiEdit integrates the temporal Mamba module with temporal-aware scanning to effectively enhance the temporal consistency of edited videos, while also demonstrating that the Mamba-based module has competitive potential in temporal modeling compared to attention mechanisms.
\item[$\bullet$] We design a noise injection strategy based on the Fourier transform to refine the initial latent noise during the editing, improving text alignment while ensuring temporal consistency and fidelity to the input video.
\item[$\bullet$] Extensive experiments demonstrate that EquiEdit performs favorably in text-guided video editing tasks, ensuring temporal consistency and good editability.
\end{itemize}

\begin{figure*}[t]
    \centering
    \captionsetup{type=figure}
    \includegraphics[width=1.0\textwidth]{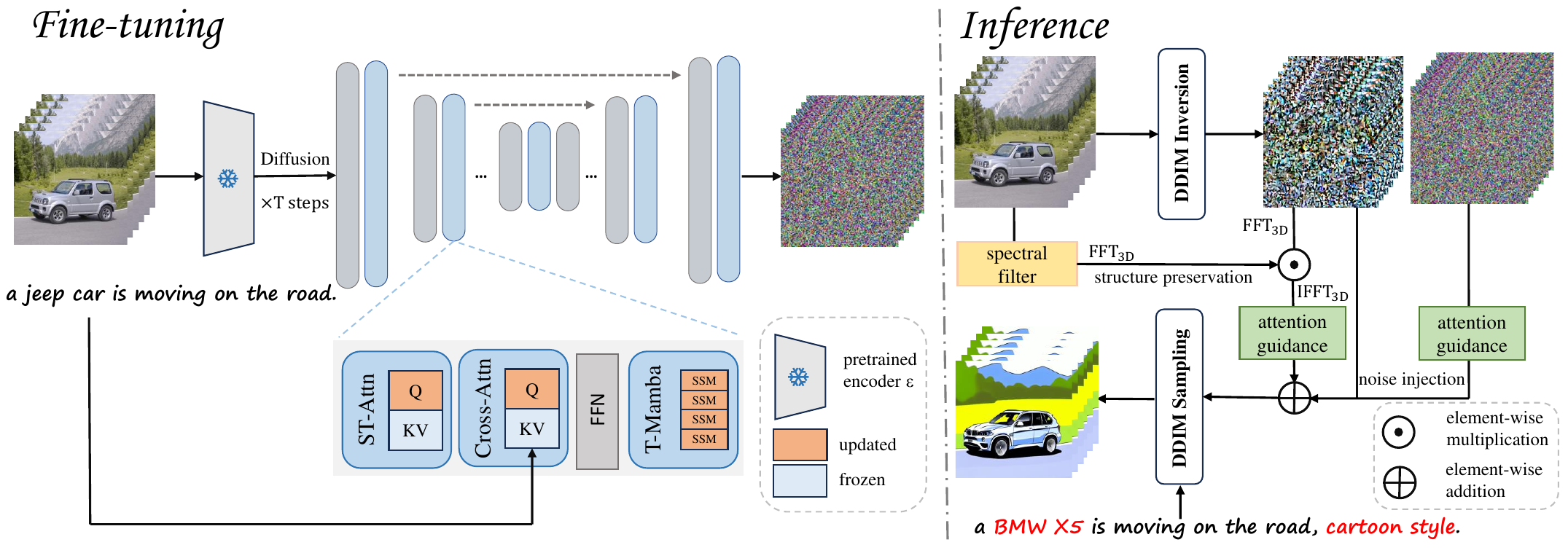}
    \caption{\textbf{An illustration of EquiEdit.} 
    Given a text-video pair, our method leverages a pre-trained T2I model for video editing. 
    During fine-tuning, we only update the projection matrices in the attention blocks and the parameters in the temporal Mamba module. During inference, we introduce Gaussian noise into the initial latent noise which is the output of inversion, while combining spectral transformations to preserve high-frequency information.
    }
    \label{fig:pdf_framework} 
\end{figure*}

\section{Related Work}
\textbf{Text-to-image (T2I) generation} based on diffusion models (DMs) has been widely studied in recent years~\cite{ruiz2023dreambooth,mou2024t2i,gu2022vector,saharia2022photorealistic,gal2022image}. Imagen~\cite{saharia2022photorealistic} effectively improves the quality of gradually generated images by introducing a cascaded diffusion model with noise conditioning augmentation~\cite{ho2022cascaded}. 
Text-inversion~\cite{gal2022image} and DreamBooth~\cite{ruiz2023dreambooth} finetunes the pre-trained T2I model with a few images and achieves binding between unique identifiers and specific themes of images provided by users for personalized T2I generation. VQ-Diffusion~\cite{gu2022vector}, similar to latent diffusion models (LDMs)~\cite{rombach2022high}, operates on the hidden space of the autoencoder for T2I generation, significantly reducing computational requirements. In our work, we utilize a pre-trained T2I model based on LDMs.
\\
\textbf{Text-guided video editing} refers to using text to guide the video editing, allowing users to express their editing requirements directly through natural language~\cite{xing2023survey}. 
Unlike image editing, video editing requires aligning the output video with text prompts while ensuring temporal consistency and fidelity to the input video's content~\cite{chai2023stablevideo,geyer2023tokenflow}. Although the methods~\cite{wang2023modelscope,chen2023control} of training a text-to-video (T2V) model for video editing on large-scale text-video pairs have good consistency, they suffer from expensive computation and time consumption. The training-free methods~\cite{ling2024motionclone,cohen2024slicedit,wang2023zero} utilize pre-trained T2I or T2V models combined with temporal modules for video editing, but still face spatiotemporal inconsistency and semantic disparity. In contrast, the few-shot or one-shot methods~\cite{wu2023tune,zhang2024towards,liu2024video,zhang2024camel} are a compromise approach that provides editing flexibility and consistency at a low training cost. 
Among them, Tune-A-Video~\cite{wu2023tune} successfully extends the pre-trained T2I model to the video domain using a tailored spatiotemporal attention and an efficient fine-tuning strategy. Video-P2P~\cite{liu2024video} converts the T2I model to a Text-to-set model (T2S) to maintain semantic consistency and proposes a decoupled-guidance strategy for improving performance. CAMEL~\cite{zhang2024camel} introduces a causal motion-enhanced attention to enhance the motion information. Although these methods have strengthened temporality, the generated videos still exhibit inconsistency across frames or shortcomings in editability.

We propose a temporal consistency enhancement module based on the SSM~\cite{gu2023mamba} and a temporal-aware scanning for long-context modeling of video. S4~\cite{gu2021efficiently} and Mamba~\cite{gu2023mamba} have demonstrated that SSMs have fast inference ability and linear scaling of sequence length, compared to the quadratic scaling of self-attention.
Thus, our module can more effectively improves finetuning efficiency than temporal attention.

\section{Methodology}
In this section, we present our proposed framework EquiEdit, the temporal Mamba module, and the noise injection strategy. The pipeline of our framework is illustrated in Fig.~\ref{fig:pdf_framework}. 
Consider a video in the latent space defined as $\mathbf{X}_{v}\in{\mathbb{R}}^{F\times W\times H\times C}$, where $F$, $W$, $H$, and $C$ represent the number of frames, width, height, and channels of each frame, respectively.
We define the given text prompt as $\mathcal{P}$.

\begin{figure*}[t]
    \centering
    \captionsetup{type=figure}
    \includegraphics[width=1.0\textwidth]{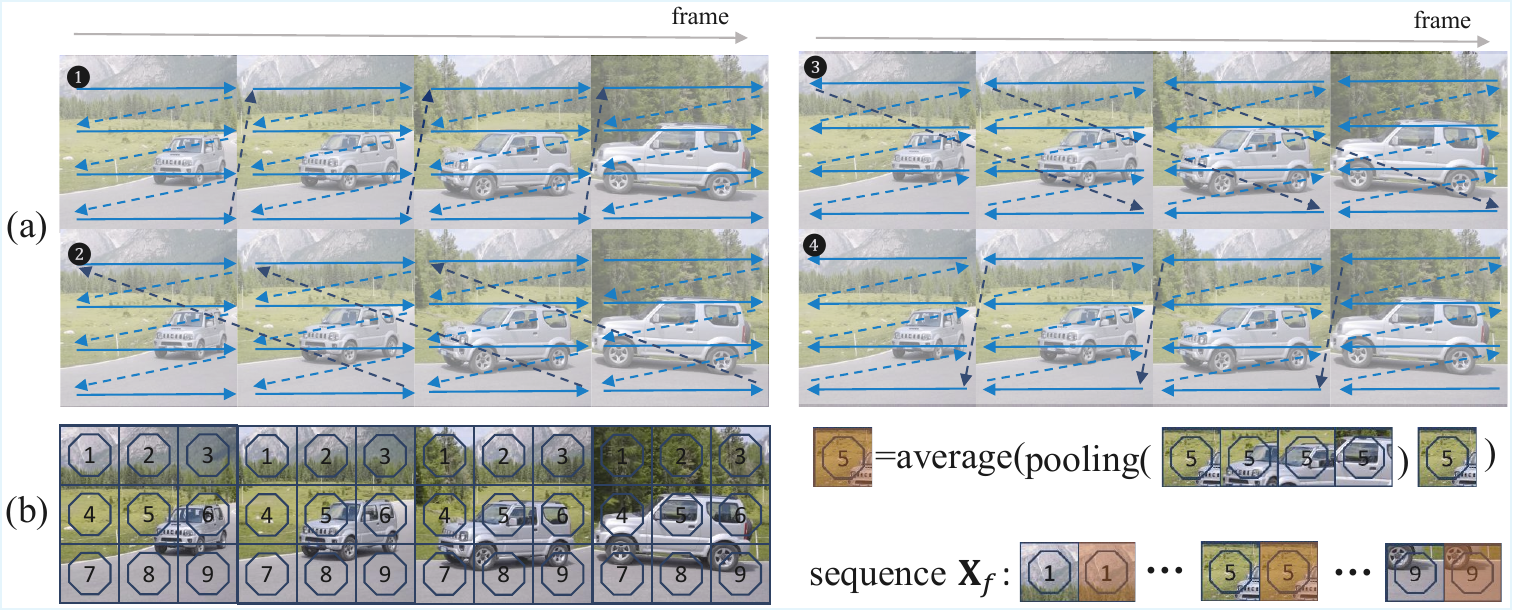}
    \caption{Visualization of the four spatial-first scanning directions (a) and temporal-aware scanning (b) in the temporal Mamba module. The image blocks represent tokens of the visual sequence $\mathbf{X}_{s}$, while the light red blocks represent tokens of the temporal sequence $\mathbf{X}_{t}$. The final fused sequence $\mathbf{X}_{f}$ is created by the token-wise concatenation of $\mathbf{X}_{s}$ and $\mathbf{X}_{t}$. Our temporal Mamba module scan the fused sequence $\mathbf{X}_{f}$ in (b) along the four directions shown in (a).}
    \label{fig:pdf_scan_driection} 
\end{figure*}

\subsection{Temporal Mamba Module} 
\label{subsec:temporal mamba module}
We propose a temporal Mamba module with a designed temporal-aware scanning to process flattened fused non-causal video sequences for better temporal modeling.
Specifically, to process 3D videos, we first convert $\mathbf{X}_{v}$ into a sequence of tokens by reshaping it, which is represented as $\mathbf{X}_{s}\in\mathbb{R}^{L \times C}$, where $L=F\times H\times W$ defines the length of the sequence, and $C$ defines the dimension of each token at this point. 
Since a video sequence is a long sequence, to avoid the shortcomings of long-context modeling capabilities, we design the temporal-aware scanning to enable the model to obtain overall temporal information to increase the receptive field.
Firstly, we apply the average pooling to $\mathbf{X}_{s}$ along the temporal dimension and further average the pooling result and tokens in $\mathbf{X}_{s}$ to create the temporal sequence $\mathbf{X}_{t}$, which helps mitigate the excessive impact of temporal information on modeling. Next, we arrange $\mathbf{X}_{s}$ and $\mathbf{X}_{t}$ token by token to form the final fused sequence $\mathbf{X}_{f}\in\mathbb{R}^{2L \times C}$, where $L=F\times H\times W$. After scanning, we only extract the $\mathbf{X}_{s}$ part that has learned long-context information. This approach allows the original visual sequence $\mathbf{X}_{s}$ to directly capture long-context information from $\mathbf{X}_{t}$ during scanning, thus enhancing the long-context modeling capability of Mamba.
Based on this, we propose our temporal Mamba module, which integrates four spatial-first SSM scanning paths in different directions to scan the fused sequence $\mathbf{X}_{f}$ and complement the spatiotemporal modeling of the video, which are illustrated in Fig.~\ref{fig:pdf_scan_driection}. More specifically, the four directions are as follows: maintaining the relative positions of tokens within frames while arranging the frames in sequential and reverse order (\ding{182} and \ding{183}), and reversing the relative order of tokens within frames while arranging the frames in order and in reverse (\ding{184} and \ding{185}). This method captures the spatial and corresponding global temporal information of tokens in various ways, thereby enhancing the receptive field and effectively improving the consistency between edited video frames. 
Compared with key-frame-based attention mechanisms~\cite{wu2023tune}, our temporal Mamba module supports full global temporal modeling. Although attention mechanisms can facilitate global temporal modeling, this is impractical due to $H \times W \gg F$, leading to excessive memory requirements that significantly increase with the increase in the frame number $F$ because of its quadratic complexity.


\begin{figure*}[t]
    \centering
    \captionsetup{type=figure}
    \includegraphics[width=0.99\textwidth]{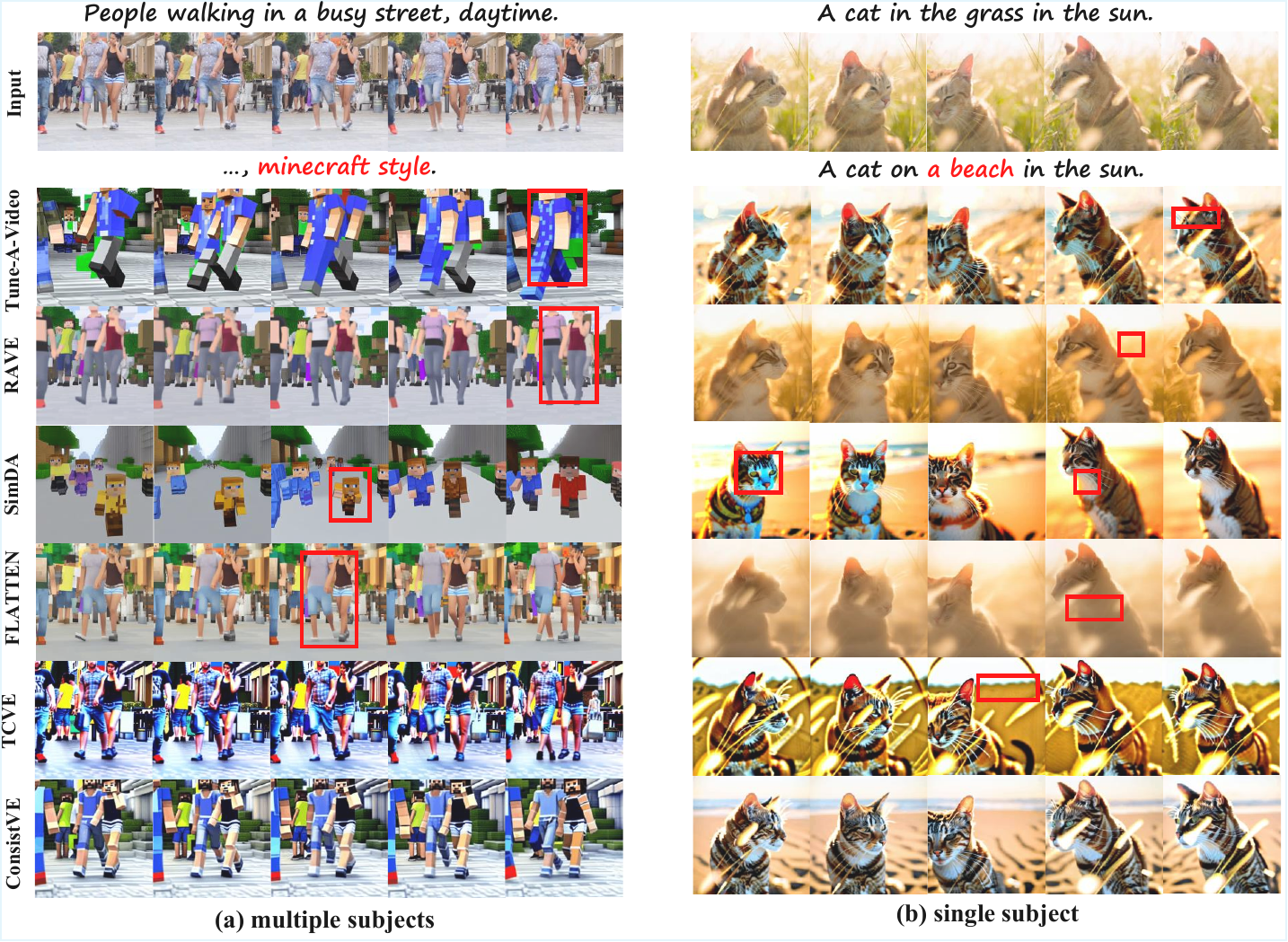}
    \caption{Qualitative comparison between evaluated methods. EquiEdit demonstrates a balance between temporal consistency and editability, remaining faithful to the input video itself while ensuring that the edits are natural.}
    \label{fig:pdf_comparsion} 
\end{figure*}

\subsection{Noise Injection Strategy}
\label{subsec:noise injection strategy}
The lack of high-frequency components leads to flickering and blurring in the editing results in the denoising process~\cite{yoon2024frag}.
Therefore, we propose a Fourier transform-based noise injection strategy, which adds editing flexibility to the initial latent noise $\mathbf{Z}_{v}$ by injecting extra noise into it. This approach also helps maintain temporal consistency by spectral modulation while avoiding flicking and blurring. We first selectively filter out the low-frequency components of the video $\mathbf{X}_{v}$ in the Fourier domain to obtain the high-frequency structural information of the input video. This process can be mathematically represented as
\begin{equation}
\begin{aligned}
    \mathbf{X}_{v}^{h} = \mathcal{IFFT}_\mathrm{3D}(\mathcal{FFT}_\mathrm{3D}(\mathbf{X}_{v})\odot \mathbf{M}),
\end{aligned}
\end{equation}
where $\mathcal{FFT}_\mathrm{3D}$ and $\mathcal{IFFT}_\mathrm{3D}$ represent the 3D Fast Fourier transform and the 3D inverse Fast Fourier transform, respectively, and $\odot$ denotes the element-wise multiplication. $\mathbf{M}$ is a Fourier mask used to implement a frequency-dependent scaling coefficient $c$,  defined as
\begin{equation}
\label{equ:mask}
\begin{aligned}
    \mathbf{M}(r)=\begin{cases}c &\mathrm{if} \ r<r_\mathrm{thr},
    \\1&\mathrm{otherwise},\end{cases}
\end{aligned}
\end{equation}
where $r$ and $r_\mathrm{thr}$ are the radius and the frequency threshold, respectively. After that, we further construct a frequency pass filter based on $\mathbf{X}_{v}^{h}$ to appropriately scale the initialized latent noise $\mathbf{Z}_{v}$ in the frequency domain for guiding the extraction of the parts of $\mathbf{Z}_{v}$ related to high-frequency visual components, which can be represented as
\begin{equation}
\begin{aligned}
    \mathbf{Z}_{v}^{h} = \mathcal{IFFT}_\mathrm{3D}(Norm(\mathcal{FFT}_\mathrm{3D}(\mathbf{X}_{v}^{h}))\odot \mathcal{FFT}_\mathrm{3D}(\mathbf{Z}_{v})),
\end{aligned}
\end{equation}
where $Norm$ is the min-max normalization. $\mathbf{Z}_{v}^{h}$ retains the part of the initial latent noise $\mathbf{Z}_{v}$ corresponding to high-frequency structural information of the input video $\mathbf{X}_{v}$, which can prevent the injection of noise from causing the lack of texture details in each edited frame, thereby may lead to inconsistency between frames. After that, based on attention guidance, we inject Gaussian noise into the initial latent noise and selectively retain the high-frequency information within it. Specifically, with the aid of the cross-attention maps between text prompts $\mathcal{P}$ and $\mathbf{X}_{v}$ in the inversion, we introduce noise into $\mathbf{Z}_{v}$. Here, we choose the cross-attention maps from the down-block and mid-block of the diffusion model, which can be represented as $\mathbf{A}_{down}$ and $\mathbf{A}_{mid}$, respectively. Consequently, the amount of noise added to the region can be expressed as
\begin{equation}
\label{equ:noise_attention}
\begin{aligned}
    \mathbf{A}_{noise} = Norm(\alpha \times \mathbf{A}_{down} + (1-\alpha) \times \mathbf{A}_{mid}),
\end{aligned}
\end{equation}
where $\alpha$ is a hyperparameter that control the ratio of $\mathbf{A}_{down}$ and $\mathbf{A}_{mid}$. Through the guidance of the cross-attention maps, we can appropriately introduce  the noise $ \epsilon$ and structural information $\mathbf{Z}_{v}^{h}$ from the initial hidden noise $\mathbf{Z}_{v}$. Here, we refer to $\mathbf{A}_{noise} \times \epsilon + (1-\mathbf{A}_{noise}) \times \mathbf{Z}_{v}^{h}$ as the additional editable information. Thus, the final latent noise can be mathematically represented as

\begin{equation}
\label{equ:final_latent_noise}
\begin{aligned}
    \mathbf{Z}^{*}_{v} = \mathbf{Z}_{v} + \gamma\times (\mathbf{A}_{noise} \times \epsilon + (1-\mathbf{A}_{noise}) \times \mathbf{Z}_{v}^{h}),
\end{aligned}
\end{equation}
where $\epsilon$ is the Gaussian noise and $\gamma$ is the hyperparameter that adjusts the ratio of the additional editable information. $\mathbf{Z}^{*}_{v}$ can enhance the editability of the video while adaptively incorporating the characteristics of $\mathbf{Z}_{v}^{h}$, ensuring that the details of the edited results do not become blurred and unfaithful to the input video. Our noise injection strategy is performed only once before sampling and introduces no trainable parameters, making its computational cost negligible.

\section{Experiments}
In this section, we qualitatively and quantitatively compare our method with SOTA methods, present an ablation study, and assess the effectiveness of our temporal-aware scanning as well as the additional editable information.

\subsection{Experimental Settings}
\noindent
\textbf{Dataset.}
We conduct experiments on $76$ videos from LOVEU-TGVE~\cite{wu2023cvpr}, including $16$ videos from DAVIS~\cite{perazzi2016benchmark}, $37$ videos from Videvo, and the remaining $23$ videos from Youtube. This is a commonly used dataset in video editing~\cite{wu2023tune,zhang2024camel,wang2024edit,cong2023flatten}. Each video comes with a ground-truth caption and $4$ text prompts, each about style transfer, object editing, background changes, and multiple changes. Each video is uniformly sampled into $32$ frames with a resolution of $512\times512$. 

\noindent
\textbf{Implementation Details.}
Our experiment is based on Latent Diffusion Models~\cite{rombach2022high} and publicly available pre-trained weight of stable diffusion v$1.4$ from the official Huggingface repo for a fair comparison. We finetune our model $500$ steps on a learning rate of $3\times10^{-5}$ and a batch size of $1$. At inference, we implement $50$ timesteps for DDIM inversion and sampling~\cite{song2020denoising} with classifier-free guidance~\cite{ho2022classifier} of magnitude $12.5$. 
The hyperparameters $c$, $r_\mathrm{thr}$, $\alpha$, and $\gamma$ in Eq. (\ref{equ:mask}), Eq. (\ref{equ:noise_attention}), and Eq. (\ref{equ:final_latent_noise}) are set to $0.2$, $1$, $0.3$, and $0.5$, respectively.
The experiments are executed on NVIDIA H$800$.

\begin{table}[t]
    \centering
    \caption{Quantitative comparison for frame consistency.}
    \resizebox{0.9\linewidth}{!}{\begin{tabular}{l|cc}
    \toprule
    \textbf{Method} & {\textbf{CLIP Score $\uparrow$}} & {\textbf{User Vote $\uparrow$}}\\
    \midrule
    Tune-A-Video~\cite{wu2023tune}   & 94.943 & 20.8$\%$\\
    SimDA~\cite{xing2024simda}     &  91.648 & 17.8$\%$\\
    RAVE~\cite{kara2024rave}     & 95.344 & 36.1$\%$\\
    FLATTEN~\cite{cong2023flatten} & 95.617  & 40.5$\%$\\
    TCVE~\cite{wang2024edit} & 94.311 & 29.7$\%$\\
    EquiEdit (Ours) & \textbf{95.863} & \textbf{79.2\%} / \textbf{82.2\%} / \textbf{63.9\%} / \textbf{59.5\%} / \textbf{70.3\%} \\
    \bottomrule
    \end{tabular}}
    \label{tab:frame_consistency}
\end{table}

\begin{table}[t]
    \centering
    \caption{Quantitative comparison for text alignment.}
    \resizebox{0.9\linewidth}{!}{\begin{tabular}{l|cc}
    \toprule
    \textbf{Method} & {\textbf{CLIP Score $\uparrow$}} & {\textbf{User Vote $\uparrow$}}\\
    \midrule
    Tune-A-Video~\cite{wu2023tune} & 30.151 &  31.6$\%$\\
    SimDA~\cite{xing2024simda}   &  30.940 & 33.6$\%$\\
    RAVE~\cite{kara2024rave}   &  30.153 &  24.7$\%$\\
    FLATTEN~\cite{cong2023flatten} & 30.832 & 26.4$\%$\\
    TCVE~\cite{wang2024edit} & 29.969 & 30.2$\%$\\
    EquiEdit (Ours) & \textbf{31.068} & \textbf{68.4\%} / \textbf{66.4\%} / \textbf{75.3\%} / \textbf{73.6\%} / \textbf{69.8\%}\\
    \bottomrule
    \end{tabular}}
    \label{tab:text_alignment}
\end{table}

\begin{table}[!t]
    \centering
    \caption{Role of temporal-aware scanning.}
    \resizebox{0.9\linewidth}{!}{\begin{tabular}{l|cc}
    \toprule
    \textbf{Method} & {\textbf{frame consistency $\uparrow$}} & {\textbf{text alignment $\uparrow$}}\\
    \midrule
    w/o temporal-aware scanning & 95.607 & 31.024 \\
    w/ temporal-aware scanning (Ours) & \textbf{95.863} & \textbf{31.068} \\
    \bottomrule
    \end{tabular}}
    \label{tab:temporal-aware-scanning}
\end{table}

\begin{figure}[t]
    \centering
    \captionsetup{type=figure}
    \includegraphics[width=0.48\textwidth]{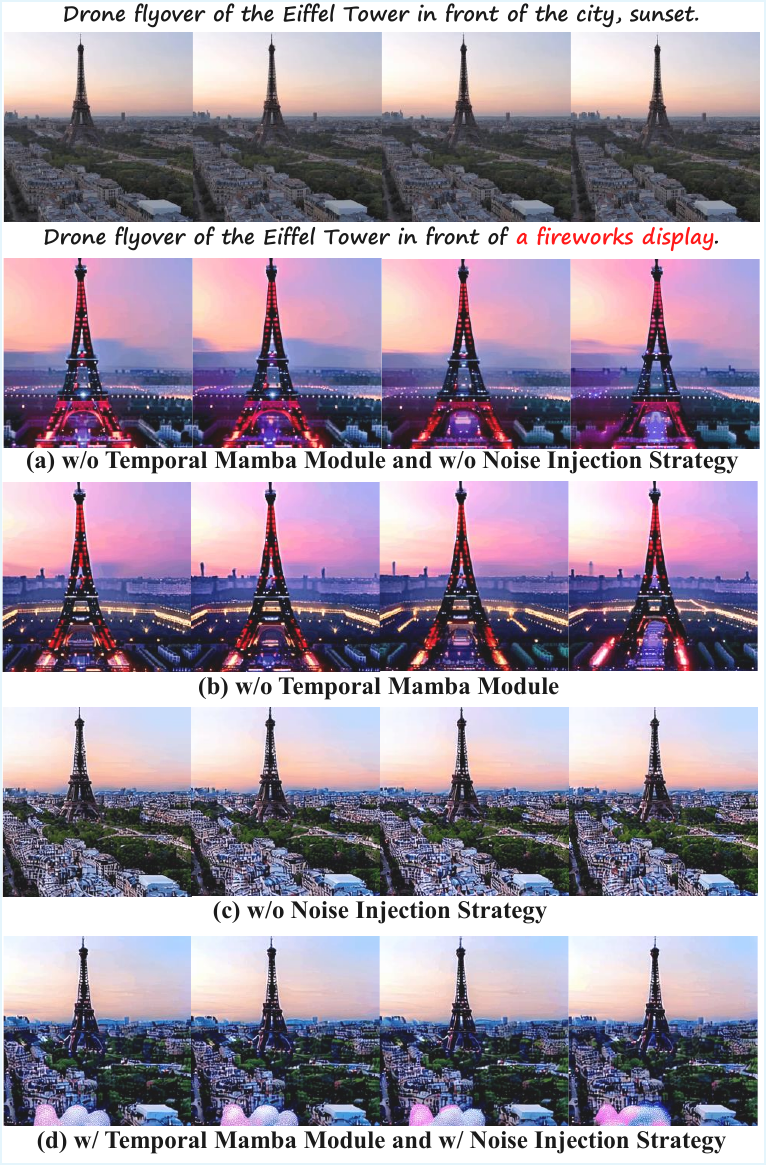}
    \caption{Ablation study of EquiEdit.}
    \label{fig:pdf_ablation} 
\end{figure}


\subsection{Evaluation}
\noindent
\textbf{Baselines.}
We compare our method with five baselines: 1) Tune-A-Video~\cite{wu2023tune}: a video editing model that is fine-tuned using a one-shot manner, which employs a temporal attention module to address the issue of temporal inconsistency. 2) RAVE~\cite{kara2024rave}: a cutting-edge video editing approach that firstly uses the grid trick for editing research. In our experiment, the grid size is set to $2\times 2$ for the $32$-frame videos. 3) SimDA~\cite{xing2024simda}: a recent parameter-efficient video diffusion model based on fine-tuning for text-guided video editing and generation. 4) FLATTEN~\cite{cong2023flatten}: A novel optical flow-guided T2V editing method. 5) TCVE~\cite{wang2024edit}: A recent method that emphasizes temporal consistency based on a T2I model. 

\noindent
\textbf{Qualitative Results.}
We present some examples of video editing in Fig.~\ref{fig:pdf_teaser}, while Fig.~\ref{fig:pdf_comparsion} showcases comparisons with the edited results of the state-of-the-art approaches. Accordingly, one can see that our method ensures visual consistency while achieving more effective editing across multiple different scenarios. In Fig.~\ref{fig:pdf_comparsion}, compared to Tune-A-Video, SimDA, FLATTEN, and TCVE, our method provides more detailed visual content, particularly in (b) where the hair details of the ``cat" are clearer. Additionally, our method exhibits a high level of alignment with editing requirements in scenes with multiple subjects, i.e., (a). In contrast to RAVE and optical flow-guided FLATTEN, our method not only excels in terms of detail but also demonstrates stronger editability, aligning more closely with the edited prompts, which demonstrates that the edits are seamless and produce coherent and natural editing results. Furthermore, our method shows fidelity to the input video in terms of motion consistency. 

\noindent
\textbf{Quantitative Results.}
We employ the CLIP score~\cite{radford2021learning} and a user study with $5$ participants to assess the frame consistency and text alignment. Finally, we analyze the effect of temporal-aware scanning.

\textit{{Frame Consistency.}}
We calculate per-frame global CLIP cosine similarities and report the mean of these values for frame consistency evaluation. We also conduct a user study to evaluate user preferences regarding frame consistency. Participants are asked to select the edited videos with the best frame consistency, with the voting rates being tallied as the final results. The results in Tab.~\ref{tab:frame_consistency} show our method achieves the best consistency score in both the CLIP score and user study, particularly outperforming the second-best method (i.e., FLATTEN) by $0.2$ in terms of consistency. 

\textit{Text Alignment.}
To assess the text alignment, the mean CLIP score between the edited videos and the editing prompts is reported. Accordingly, in the user study, participants are asked to select the results with the best text alignment. The results in Tab.~\ref{tab:text_alignment} show that our method also achieves the best editability. Additionally, by comprehensive analysis of Tab.~\ref{tab:frame_consistency} and Tab.~\ref{tab:text_alignment}, while competing methods show a trade-off between frame consistency and editability, our method achieves a balance between the two and demonstrates higher user preference in practical scenarios.

\textit{Effectiveness of Temporal-aware Scanning.} We removed temporal-aware scanning in the proposed temporal Mamba module, similar to most methods~\cite{zhu2024vision,li2024videomamba}, allowing our module to directly scan the visual sequence $\mathbf{X}_{s}$ along the four directions in Fig.~\ref{fig:pdf_scan_driection} (a), with results in Tab.~\ref{tab:temporal-aware-scanning}. This result indicates that the designed temporal-aware scanning effectively enhances temporal consistency and that scanning the fused sequence $\mathbf{X}_{f}$ mitigates Mamba's limitations in long-context modeling capability.


\subsection{Ablation Study}
We perform an ablation study to assess the core components of our model in Fig.~\ref{fig:pdf_ablation}: the temporal Mamba module and the noise injection strategy. 
The variants ``w/o temporal Mamba module and w/o noise injection strategy" (a) and ``w/o temporal Mamba module" (b) are created by removing both components and only the temporal Mamba module, respectively.
The ``w/o noise injection strategy" variant (c) is conducted by removing the noise injection strategy from our model. Finally, we present the results of our model in (d). The results in Fig.~\ref{fig:pdf_ablation} demonstrate that the temporal Mamba module plays a crucial role in enhancing the fidelity and consistency of the editing results to the input video. Without the temporal Mamba module, the editing results are primarily influenced by the knowledge from the pre-trained parameters, exhibiting significant infidelity and semantic disparity. By combining the temporal Mamba module and the noise injection strategy, our model achieves an improvement and balance in consistency and editability, while maintaining the fidelity to the input video. 

\begin{figure}[t]
    \centering
    \captionsetup{type=figure}
    \includegraphics[width=0.48\textwidth]{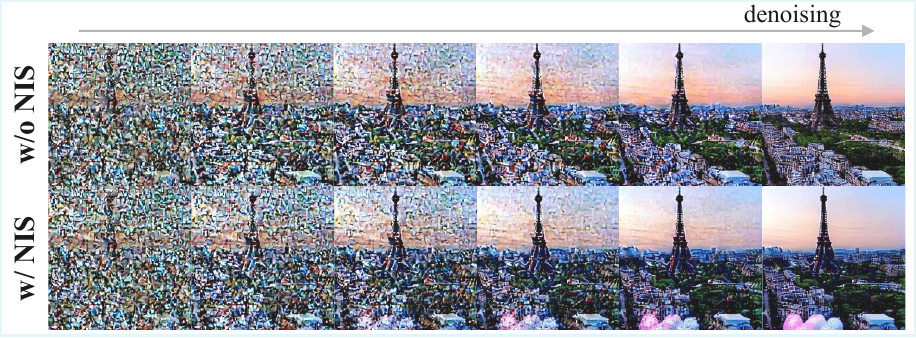}
    \caption{Visualization of the corresponding denoising process. NIS refers to the noise injection strategy.}
    \label{fig:pdf_denoising} 
\end{figure}

To further illustrate the impact of the noise injection strategy on the diffusion sampling process, Fig.~\ref{fig:pdf_denoising} presents the visualization of the corresponding denoising processes. We can see that the noise injection strategy successfully influences the denoising process, enhancing the editability while avoiding blurring.

\subsection{Impact of Additional Editable Information}
\begin{figure}[t]
    \centering
    \captionsetup{type=figure}
    \includegraphics[width=0.48\textwidth]{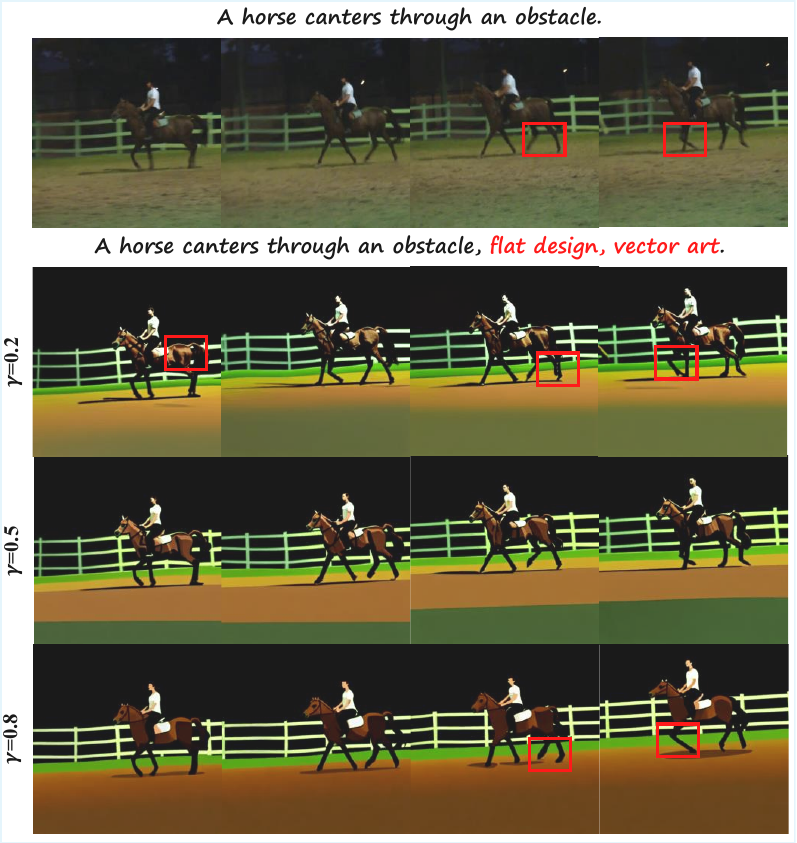}
    \caption{The impact of additional editable information. 
    As $\gamma$ rises, editing results better match flat vector design requirements, yet visuals become rigid and structurally inconsistent.
    }
    \label{fig:pdf_additional_editable} 
\end{figure}

We conduct further experiments to demonstrate the impact of additional editable information, specifically the influence of noise $\epsilon$ and structural information $\mathbf{Z}_{v}^{h}$ guided by the cross-attention maps on the editing effects. 
We tested various values of its hyperparameter $\gamma$, and the experimental results are shown in Fig.~\ref{fig:pdf_additional_editable}.
We can observe that when $\gamma$ is set too low (i.e., $\gamma = 0.2$), the editing results do not align well with the requirements of "flat design, vector art" (as seen in the horse's skin), and there is also an inconsistency in the visual structure compared to the original video (notably in the horse's legs). Conversely, when $\gamma$ is set too high (i.e., $\gamma = 0.8$), although editability is achieved, the overall visuals become stiff and unnatural, with misaligned details. This indicates that the amount of injected additional editable information needs to be carefully regulated, and it also suggests that  $\gamma$ setting of $0.5$ in EquiEdit is appropriate.

\section{Conclusion}
In this paper, we present EquiEdit, a novel customizable video editing framework aimed at collaboratively enhancing the temporal consistency and editability of video. Moreover, it demonstrates a greater fidelity to the input video itself and the editing is more natural and seamless. We are the first to apply Mamba in the field of video editing to improve the consistency, showing Mamba's potential in extracting temporal information. The proposed noise injection strategy appropriately enhances editability while maintaining temporal consistency. Extensive experiments have demonstrated that our model balances consistency and editability in text-to-video editing benchmarks.

\clearpage
\bibliography{aaai2026}

\end{document}